\documentclass[11pt]{article}

\usepackage[utf8]{inputenc}
\usepackage[T1]{fontenc}
\usepackage{amsmath,amssymb,amsfonts}
\usepackage{graphicx}
\usepackage{booktabs}
\usepackage{hyperref}
\usepackage{xcolor}
\usepackage{natbib}
\usepackage[margin=1in]{geometry}

\hypersetup{
    colorlinks=true,
    linkcolor=blue,
    citecolor=blue,
    urlcolor=blue
}

\newcommand{\modelname}[1]{\textsf{#1}}

\title{The Dunning-Kruger Effect in Large Language Models:\\
An Empirical Study of Confidence Calibration}

\author{
    Sudipta Ghosh\thanks{Corresponding author.} \\
    Cognizant Technology Solutions
    \and
    Mrityunjoy Panday \\
    Cognizant Technology Solutions
}

\date{}

\begin{document}

\maketitle

\begin{abstract}
Large language models (LLMs) have demonstrated remarkable capabilities across diverse tasks, yet their ability to accurately assess their own confidence remains poorly understood. We present an empirical study investigating whether LLMs exhibit patterns reminiscent of the Dunning-Kruger effect---a cognitive bias where individuals with limited competence tend to overestimate their abilities. We evaluate four state-of-the-art models (\modelname{Claude Haiku 4.5}, \modelname{Gemini 2.5 Pro}, \modelname{Gemini 2.5 Flash}, and \modelname{Kimi K2}) across four benchmark datasets totaling 24,000 experimental trials. Our results reveal striking calibration differences: \modelname{Kimi K2} exhibits severe overconfidence with an Expected Calibration Error (ECE) of 0.726 despite only 23.3\% accuracy, while \modelname{Claude Haiku 4.5} achieves the best calibration (ECE = 0.122) with 75.4\% accuracy. These findings demonstrate that poorly performing models display markedly higher overconfidence---a pattern analogous to the Dunning-Kruger effect in human cognition. We discuss implications for safe deployment of LLMs in high-stakes applications.
\end{abstract}


\section{Introduction}
\label{sec:introduction}

The Dunning-Kruger effect, first described by Kruger and Dunning~\citep{dunning1999unskilled}, refers to a cognitive bias in which individuals with limited knowledge or competence in a domain tend to overestimate their own abilities, while those with greater expertise may provide more accurate self-assessments. Recent work has begun exploring whether LLMs exhibit analogous confidence-competence gaps~\citep{singh2024confidence,singh2023cognitive}, revealing systematic patterns of miscalibration that mirror human cognitive biases. As large language models become increasingly integrated into decision-making systems, understanding whether these models exhibit similar patterns of miscalibration becomes critical for safe deployment~\citep{steyvers2024llms}.

Recent advances in LLM capabilities have been accompanied by growing concerns about their reliability and trustworthiness~\citep{bommasani2021opportunities}. A model that expresses high confidence in incorrect answers poses significant risks, particularly in high-stakes domains such as healthcare, legal reasoning, and scientific research. Conversely, a model that underestimates its confidence in accurate responses may fail to provide valuable insights when they are most needed.

This study addresses the following research questions:
\begin{enumerate}
    \item Do LLMs exhibit systematic overconfidence in their responses, and does the magnitude of miscalibration correlate inversely with task performance?
    \item How does confidence calibration vary across different model families, architectures, and knowledge domains?
    \item Can we identify patterns analogous to the Dunning-Kruger effect in LLM behavior, where lower-performing models exhibit greater overconfidence?
\end{enumerate}

Our contributions are as follows:
\begin{itemize}
    \item We conduct a comprehensive empirical study of confidence calibration across four LLMs and four benchmarks, totaling 24,000 experimental trials.
    \item We provide quantitative evidence that poorly performing models exhibit disproportionate overconfidence, mirroring the Dunning-Kruger effect.
    \item We identify \modelname{Claude Haiku 4.5} as exhibiting superior calibration properties, including appropriate underconfidence in uncertain domains.
    \item We release our experimental framework and analysis pipeline for reproducibility.
\end{itemize}

\section{Related Work}
\label{sec:related}

\subsection{Confidence Calibration in Machine Learning}

Calibration in machine learning refers to the alignment between predicted confidence and actual accuracy~\citep{guo2017calibration}. A well-calibrated model should express $p\%$ confidence in predictions that are correct approximately $p\%$ of the time. Guo et al.~\citep{guo2017calibration} demonstrated that modern neural networks are often miscalibrated, typically exhibiting overconfidence. Naeini et al.~\citep{naeini2015obtaining} introduced Expected Calibration Error (ECE) as a principled metric for quantifying miscalibration.

\subsection{Confidence-Competence Gap in LLMs}

Recent research has investigated whether LLMs exhibit human-like confidence-competence gaps. Singh et al.~\citep{singh2024confidence,singh2023cognitive} conducted cognitive studies revealing systematic overconfidence patterns in LLMs that parallel the Dunning-Kruger effect. This line of work has been extended to specialized domains, with Singh et al.~\citep{singh2025code} demonstrating similar effects in code generation models. Xu et al.~\citep{xu2025mirror} explored psychological insights from human confidence research to understand and address overconfidence in LLMs, while Steyvers et al.~\citep{steyvers2024llms} examined the gap between what LLMs actually know and what users perceive them to know.

\subsection{Uncertainty Quantification in LLMs}

Recent work has explored various approaches to uncertainty quantification in language models. Kadavath et al.~\citep{kadavath2022language} found that language models ``mostly know what they know,'' exhibiting some ability to express uncertainty through verbalized confidence. Park et al.~\citep{park2026finetuning} further proposed fine-tuning methods to improve metacognitive alignment, enabling models to more accurately report what they actually know. Lin et al.~\citep{lin2022teaching} proposed methods for teaching models to express uncertainty in natural language. Kuhn et al.~\citep{kuhn2023semantic} introduced semantic uncertainty as a more robust measure that accounts for linguistic variation. Xiong et al.~\citep{xiong2023uncertainty} provided an empirical evaluation of confidence elicitation methods in LLMs, finding that verbalized confidence often diverges from actual performance. For comprehensive surveys, see Geng et al.~\citep{geng2024naacl} and Xie et al.~\citep{xie2024survey}.

\subsection{Overconfidence and Calibration Methods}

LLM overconfidence has emerged as a critical concern. Groot and Valdenegro-Toro~\citep{groot2024overconfidence} demonstrated that verbalized uncertainty in LLMs is systematically overconfident. Chhikara~\citep{chhikara2025confidence} examined how distractor effects influence calibration, while Leng et al.~\citep{leng2024taming} proposed reward calibration during RLHF to mitigate overconfidence. Zhu et al.~\citep{zhu2023calibration} investigated the relationship between model alignment and calibration quality. The connection between miscalibration and hallucination has also been explored~\citep{zhang2023hallucination}, suggesting shared underlying mechanisms. Recent work on reasoning models indicates that extended thinking may improve confidence calibration~\citep{yoon2025reasoning,pawitan2024confidence}.

\subsection{LLM Benchmarking}

Standardized benchmarks have become essential for evaluating LLM capabilities. MMLU~\citep{hendrycks2021measuring} tests knowledge across 57 subjects, TriviaQA~\citep{joshi2017triviaqa} evaluates factual recall through open-ended questions, and ARC~\citep{clark2018think} assesses scientific reasoning. These benchmarks span diverse cognitive demands and enable systematic cross-model comparison.

\section{Methodology}
\label{sec:methodology}

\subsection{Experimental Design}

We conducted an empirical study to investigate confidence calibration in LLMs, specifically examining whether these models exhibit Dunning-Kruger-like behavior. Our experimental design employed a factorial structure crossing four state-of-the-art LLMs with four established benchmark datasets.

All experiments were conducted with extended thinking mode enabled, using a token budget of 8,192 tokens for the reasoning process. This configuration was chosen to evaluate model calibration under conditions that maximize reasoning capability.

\subsection{Models Under Evaluation}

We evaluated four large language models representing different architectural approaches:

\begin{enumerate}
    \item \textbf{\modelname{Claude Haiku 4.5}} (Anthropic): A compact model optimized for efficiency while maintaining strong reasoning capabilities.
    \item \textbf{\modelname{Gemini 2.5 Pro}} (Google): Google's flagship reasoning model with advanced multi-step problem-solving capabilities.
    \item \textbf{\modelname{Gemini 2.5 Flash}} (Google): A faster variant designed for rapid inference while preserving accuracy.
    \item \textbf{\modelname{Kimi K2}} (Moonshot AI): A reasoning-focused model with extended thinking capabilities.
\end{enumerate}

All models were accessed through their respective APIs with temperature set to 0.0 to ensure deterministic outputs and maximize reproducibility.

\subsection{Benchmark Datasets}

We selected four widely-used benchmark datasets:

\textbf{MMLU}~\citep{hendrycks2021measuring}: A comprehensive benchmark comprising 14,042 multiple-choice questions across 57 subjects organized into STEM, humanities, social sciences, and other domains.

\textbf{TriviaQA}~\citep{joshi2017triviaqa}: A reading comprehension dataset containing 95,956 question-answer pairs requiring free-form text responses.

\textbf{ARC}~\citep{clark2018think}: A science question benchmark partitioned into Easy and Challenge subsets, with Challenge questions requiring genuine reasoning.

\textbf{HellaSwag}: A commonsense reasoning benchmark with 10,042 sentence completion tasks, adversarially filtered to be challenging for language models.

\subsection{Sample Size}

We sampled 1,500 questions per model per dataset, yielding 6,000 questions per model and 24,000 total experimental trials. This sample size provides statistical power exceeding 0.99 for detecting medium effect sizes ($d = 0.5$) at $\alpha = 0.05$, while remaining computationally feasible given API rate constraints. Questions were randomly sampled using a fixed random seed (42) for reproducibility, with stratification by subject category for MMLU, difficulty level for ARC, source for TriviaQA, and activity domain for HellaSwag.

\subsection{Confidence Elicitation Protocol}

We employed a numeric scale elicitation method, prompting models to provide both an answer and a confidence score on a 0--100 scale:

\begin{quote}
\textit{Answer the following question and rate your confidence on a scale from 0 to 100, where 0 means completely uncertain and 100 means absolutely certain.}
\end{quote}

\subsection{Evaluation Metrics}

\textbf{Expected Calibration Error (ECE)}: The primary calibration metric, computed as:
\begin{equation}
    \text{ECE} = \sum_{b=1}^{B} \frac{n_b}{N} |acc_b - conf_b|
\end{equation}
where $B = 10$ bins partition the confidence range, $n_b$ is the sample count in bin $b$, $acc_b$ is accuracy within bin $b$, and $conf_b$ is mean confidence within bin $b$.

\textbf{Correlation Coefficients}: Pearson and Spearman correlations measure the relationship between confidence scores and correctness.

\textbf{Overconfidence Score}: Defined as $\text{mean\_confidence} - \text{accuracy}$, measuring systematic bias direction.

\section{Results}
\label{sec:results}

\subsection{Overall Model Performance}

Table~\ref{tab:summary} presents summary statistics for all evaluated models. Performance varied substantially, with \modelname{Gemini 2.5 Pro} achieving the highest accuracy (80.9\%) and \modelname{Kimi K2} the lowest (23.3\%).

\begin{table}[t]
\caption{Summary statistics across all models. Accuracy and ECE are reported with 95\% bootstrap confidence intervals.}
\label{tab:summary}
\centering
\begin{tabular}{@{}lcccc@{}}
\toprule
\textbf{Model} & \textbf{Accuracy (\%)} & \textbf{Mean Conf.} & \textbf{Conf. Std.} & \textbf{ECE} \\
\midrule
Gemini 2.5 Pro & 80.9 [79.9, 81.9] & 99.5 & 16.1 & 0.185 \\
Claude Haiku 4.5 & 75.4 [74.3, 76.5] & 86.0 & 41.0 & \textbf{0.122} \\
Gemini 2.5 Flash & 70.9 [69.7, 72.0] & 97.9 & 6.8 & 0.272 \\
Kimi K2 & 23.3 [22.3, 24.4] & 95.7 & 9.4 & 0.726 \\
\bottomrule
\end{tabular}
\end{table}

\subsection{Calibration Analysis}

The most striking finding is the stark contrast in calibration quality across models. \modelname{Kimi K2} exhibits severe overconfidence, expressing mean confidence of 95.7\% despite achieving only 23.3\% accuracy, resulting in an ECE of 0.726. This represents a miscalibration gap of over 72 percentage points.

\begin{figure}[t]
\centering
\includegraphics[width=0.9\textwidth]{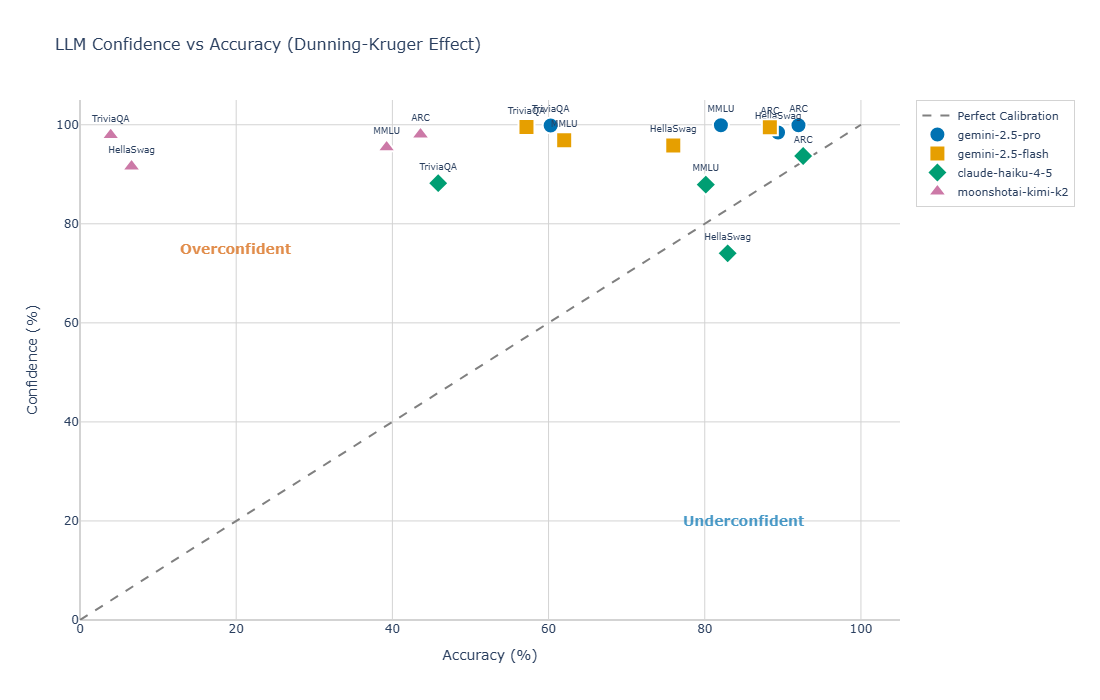}
\caption{Confidence vs. accuracy across models, demonstrating the Dunning-Kruger effect. The diagonal line represents perfect calibration. \modelname{Kimi K2} shows severe overconfidence (top-left quadrant) while \modelname{Claude Haiku 4.5} exhibits near-optimal calibration.}
\label{fig:dk}
\end{figure}

In contrast, \modelname{Claude Haiku 4.5} achieves the best calibration with ECE = 0.122. Notably, this model exhibits the highest confidence variability (std = 41.0), indicating appropriate modulation of confidence based on question difficulty. Figure~\ref{fig:dk} illustrates this Dunning-Kruger-like pattern.

\subsection{Key Findings}

Our analysis reveals several notable extremes in calibration behavior:

\begin{itemize}
    \item \textbf{Best calibration:} \modelname{Claude Haiku 4.5} on ARC achieves ECE = 0.026, representing near-perfect alignment between confidence and accuracy.
    \item \textbf{Worst calibration:} \modelname{Kimi K2} on TriviaQA exhibits ECE = 0.940, expressing 97.9\% confidence while achieving only 3.9\% accuracy---a 94 percentage point miscalibration gap.
    \item \textbf{Unique underconfidence:} \modelname{Claude Haiku 4.5} on HellaSwag is the only model-dataset combination showing underconfidence (overconfidence score = $-0.089$), with 74.0\% mean confidence despite 82.9\% accuracy.
    \item \textbf{Gemini rigidity:} Both Gemini models maintain consistently high confidence (95--99\%) regardless of actual performance, suggesting limited metacognitive modulation.
\end{itemize}

\subsection{Statistical Comparison}

One-way ANOVA revealed highly significant differences in accuracy across models ($F = 2324.7$, $p < 0.001$). Table~\ref{tab:pairwise} presents pairwise comparisons with effect sizes.

\begin{table}[t]
\caption{Pairwise model comparisons. All comparisons are statistically significant ($p < 0.05$).}
\label{tab:pairwise}
\centering
\begin{tabular}{@{}llccc@{}}
\toprule
\textbf{Model 1} & \textbf{Model 2} & \textbf{$t$-statistic} & \textbf{Cohen's $d$} & \textbf{Effect Size} \\
\midrule
Kimi K2 & Gemini 2.5 Pro & $-77.26$ & $-1.41$ & Large \\
Kimi K2 & Gemini 2.5 Flash & $-59.30$ & $-1.08$ & Large \\
Kimi K2 & Claude Haiku 4.5 & $-66.77$ & $-1.22$ & Large \\
Gemini 2.5 Pro & Gemini 2.5 Flash & $12.97$ & $0.24$ & Small \\
Gemini 2.5 Pro & Claude Haiku 4.5 & $7.36$ & $0.13$ & Negligible \\
Gemini 2.5 Flash & Claude Haiku 4.5 & $-5.58$ & $-0.10$ & Negligible \\
\bottomrule
\end{tabular}
\end{table}

\subsection{Correlation Between Confidence and Correctness}

Table~\ref{tab:correlation} presents correlation analyses. Most models show statistically significant positive correlations ($p < 0.001$), but magnitudes are weak. \modelname{Gemini 2.5 Flash} exhibits the strongest correlation ($r = 0.175$), while \modelname{Gemini 2.5 Pro}'s correlation is notably \textit{not} statistically significant ($p = 0.406$), suggesting its confidence is essentially decoupled from correctness despite achieving the highest accuracy.

\begin{table}[t]
\caption{Correlation between confidence and correctness.}
\label{tab:correlation}
\centering
\begin{tabular}{@{}lcccc@{}}
\toprule
\textbf{Model} & \textbf{Pearson $r$} & \textbf{$p$-value} & \textbf{Spearman $\rho$} & \textbf{$p$-value} \\
\midrule
Kimi K2 & 0.104 & $<0.001$ & 0.163 & $<0.001$ \\
Gemini 2.5 Pro & 0.011 & 0.406 & 0.090 & $<0.001$ \\
Gemini 2.5 Flash & 0.175 & $<0.001$ & 0.173 & $<0.001$ \\
Claude Haiku 4.5 & 0.140 & $<0.001$ & 0.167 & $<0.001$ \\
\bottomrule
\end{tabular}
\end{table}

\subsection{Per-Dataset Analysis}

\begin{figure}[t]
\centering
\includegraphics[width=0.85\textwidth]{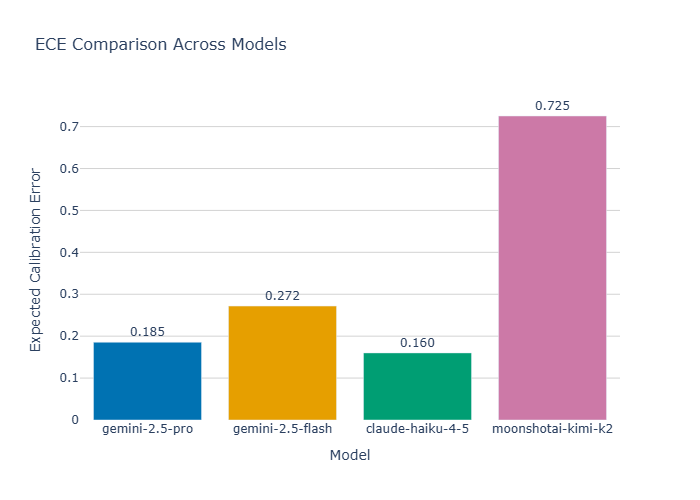}
\caption{ECE comparison across models. Lower values indicate better calibration. \modelname{Claude Haiku 4.5} achieves the best calibration, while \modelname{Kimi K2} shows severe miscalibration.}
\label{fig:ece_comparison}
\end{figure}

Figure~\ref{fig:ece_comparison} and Table~\ref{tab:dataset} reveal substantial variation across benchmarks. All models struggled most with TriviaQA (open-ended factual recall), where \modelname{Kimi K2} achieved the most extreme miscalibration observed in our study (ECE = 0.940). Performance was generally highest on ARC (scientific reasoning), where \modelname{Claude Haiku 4.5} achieved exceptional calibration (ECE = 0.026). Notably, \modelname{Claude Haiku 4.5} on HellaSwag represents the only underconfident case, with mean confidence (74.0\%) falling below accuracy (82.9\%).

\begin{table}[t]
\caption{Performance breakdown by dataset. Values are accuracy (\%) / mean confidence (\%) / ECE.}
\label{tab:dataset}
\centering
\footnotesize
\begin{tabular}{@{}lcccc@{}}
\toprule
\textbf{Dataset} & \textbf{Gemini Pro} & \textbf{Claude Haiku} & \textbf{Gemini Flash} & \textbf{Kimi K2} \\
\midrule
MMLU & 82.1 / 99.9 / 0.17 & 80.1 / 87.9 / 0.09 & 62.0 / 96.9 / 0.35 & 39.3 / 95.4 / 0.57 \\
ARC & 92.0 / 99.9 / 0.08 & 92.6 / 93.7 / \textbf{0.03} & 88.3 / 99.5 / 0.11 & 43.6 / 98.1 / 0.55 \\
HellaSwag & 89.4 / 98.4 / 0.10 & 82.9 / 74.0 / 0.10 & 76.0 / 95.8 / 0.20 & 6.6 / 91.6 / 0.85 \\
TriviaQA & 60.3 / 99.9 / 0.40 & 45.9 / 88.2 / 0.43 & 57.2 / 99.5 / 0.42 & 3.9 / 97.9 / 0.94 \\
\bottomrule
\end{tabular}
\end{table}

\subsection{Reliability Diagrams}

\begin{figure}[t]
\centering
\begin{minipage}{0.48\textwidth}
    \centering
    \includegraphics[width=\textwidth]{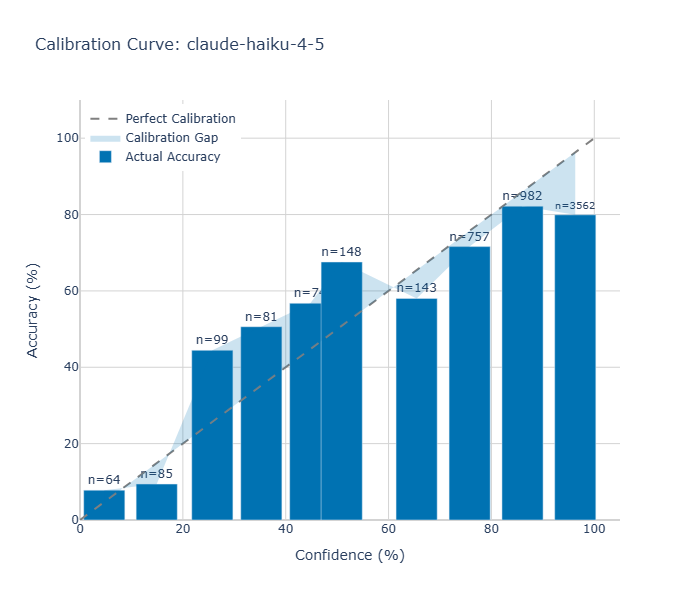}
\end{minipage}
\hfill
\begin{minipage}{0.48\textwidth}
    \centering
    \includegraphics[width=\textwidth]{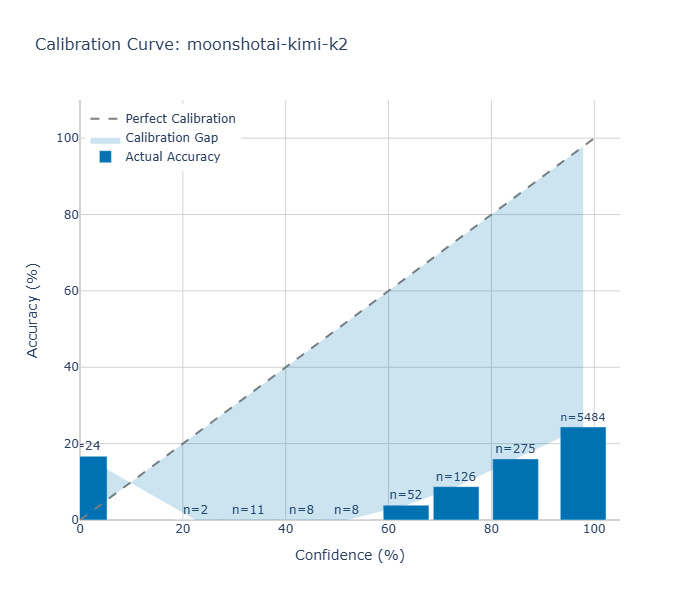}
\end{minipage}
\caption{Reliability diagrams for \modelname{Claude Haiku 4.5} (left) and \modelname{Kimi K2} (right). The diagonal represents perfect calibration. \modelname{Claude Haiku 4.5} shows well-distributed confidence with close adherence to the diagonal, while \modelname{Kimi K2} shows severe overconfidence with most responses clustered at high confidence despite low accuracy.}
\label{fig:calibration_curves}
\end{figure}

Figure~\ref{fig:calibration_curves} presents reliability diagrams for each model. \modelname{Claude Haiku 4.5} shows the closest adherence to the diagonal (perfect calibration), while \modelname{Kimi K2} shows consistent and severe deviation below the diagonal across all confidence levels.

\section{Discussion}
\label{sec:discussion}

\subsection{Evidence for Dunning-Kruger-like Patterns}

Our results provide compelling evidence for Dunning-Kruger-like patterns in LLM confidence calibration, extending prior work on confidence-competence gaps~\citep{singh2024confidence,singh2023cognitive}. The most poorly performing model (\modelname{Kimi K2}, 23.3\% accuracy) exhibited the most severe overconfidence (ECE = 0.726), while better-performing models showed progressively better calibration. This inverse relationship between competence and overconfidence mirrors the classic Dunning-Kruger effect in human cognition and aligns with observations of systematic overconfidence in LLMs~\citep{groot2024overconfidence}.

Critically, \modelname{Claude Haiku 4.5} demonstrates that good calibration is achievable. This model shows the highest confidence variability (std = 41.0), suggesting it appropriately modulates confidence based on question difficulty. Remarkably, it exhibits underconfidence on HellaSwag (overconfidence score = $-0.089$), indicating metacognitive awareness analogous to expert humility in human studies---the only such case observed across all model-dataset combinations.

\subsection{Mechanisms of Miscalibration}

The severe miscalibration observed in \modelname{Kimi K2} likely stems from training dynamics that reward confident responses regardless of correctness~\citep{leng2024taming}. Models may learn to produce high confidence scores because such responses are perceived as more helpful or authoritative during training. Without explicit calibration objectives, this bias can become severe and may contribute to hallucination tendencies~\citep{zhang2023hallucination}.

The superior calibration of \modelname{Claude Haiku 4.5} may reflect Anthropic's training approach, which emphasizes honest uncertainty expression. Zhu et al.~\citep{zhu2023calibration} found that alignment procedures significantly impact calibration quality, suggesting that calibration is substantially influenced by training methodology, not merely model scale or architecture.

\subsection{Implications for LLM Deployment}

These findings have significant implications for deploying LLMs in high-stakes applications, particularly given the gap between what LLMs know and what users perceive them to know~\citep{steyvers2024llms}:

\begin{enumerate}
    \item \textbf{Risk Assessment}: Models exhibiting Dunning-Kruger patterns are particularly dangerous because they express high confidence precisely when they are most likely to be wrong.
    \item \textbf{Model Selection}: Calibration quality should be a primary criterion for model selection, alongside raw performance metrics.
    \item \textbf{Confidence Thresholding}: Simple confidence thresholds are insufficient for poorly calibrated models; a 90\% confidence threshold provides no safety guarantee when the model's actual accuracy at that confidence level is only 24\%.
    \item \textbf{Reasoning Mode Benefits}: Our evaluation of extended thinking mode aligns with findings that reasoning models may exhibit improved calibration~\citep{yoon2025reasoning,pawitan2024confidence}.
    \item \textbf{Benchmark Limitations}: Current benchmarks that report only accuracy metrics without confidence calibration create a false belief of model competence. A model achieving 80\% accuracy may appear competent, yet express 99\% confidence on both correct and incorrect answers---masking critical reliability issues.
\end{enumerate}

\subsection{Limitations}

Several limitations should be acknowledged. First, our confidence elicitation relied on explicit prompting, which may not capture implicit model uncertainty. Second, we evaluated extended thinking mode exclusively; calibration patterns may differ under standard inference. Third, our study focused on factual and reasoning tasks; calibration in creative or open-ended tasks may exhibit different patterns. Finally, model behavior may change with API updates, limiting long-term reproducibility.

\section{Conclusion}
\label{sec:conclusion}

We presented a comprehensive empirical study of confidence calibration in large language models, revealing striking evidence for Dunning-Kruger-like patterns. Our key findings include:

\begin{enumerate}
    \item \modelname{Kimi K2} exhibits severe overconfidence (ECE = 0.726) despite achieving only 23.3\% accuracy, representing a miscalibration gap exceeding 72 percentage points.
    \item \modelname{Claude Haiku 4.5} achieves the best calibration (ECE = 0.122) with appropriate confidence modulation across difficulty levels.
    \item The inverse relationship between model competence and overconfidence provides empirical support for Dunning-Kruger-like effects in artificial systems.
    \item Calibration quality varies substantially across knowledge domains, with open-ended factual recall (TriviaQA) proving most challenging for all models.
\end{enumerate}

These findings underscore the importance of calibration evaluation as a complement to accuracy metrics when assessing LLM readiness for deployment. Future work should investigate calibration-aware training objectives and develop real-time calibration monitoring for production systems.

\section*{Acknowledgments}

This research was conducted using publicly available API access to the evaluated models. We thank the anonymous reviewers for their constructive feedback.

\bibliographystyle{plainnat}
\bibliography{references}

@article{dunning1999unskilled,
  author    = {Kruger, Justin and Dunning, David},
  title     = {Unskilled and Unaware of It: How Difficulties in Recognizing One's Own Incompetence Lead to Inflated Self-Assessments},
  journal   = {Journal of Personality and Social Psychology},
  volume    = {77},
  number    = {6},
  pages     = {1121--1134},
  year      = {1999},
  publisher = {American Psychological Association}
}

@article{bommasani2021opportunities,
  author  = {Bommasani, Rishi and Hudson, Drew A. and Adeli, Ehsan and Altman, Russ and Arora, Simran and von Arx, Sydney and Bernstein, Michael S. and Bohg, Jeannette and Bosselut, Antoine and Brunskill, Emma and others},
  title   = {On the Opportunities and Risks of Foundation Models},
  journal = {arXiv preprint arXiv:2108.07258},
  year    = {2021}
}

@inproceedings{guo2017calibration,
  author    = {Guo, Chuan and Pleiss, Geoff and Sun, Yu and Weinberger, Kilian Q.},
  title     = {On Calibration of Modern Neural Networks},
  booktitle = {Proceedings of the 34th International Conference on Machine Learning (ICML)},
  pages     = {1321--1330},
  year      = {2017},
  publisher = {PMLR}
}

@inproceedings{hendrycks2021measuring,
  author    = {Hendrycks, Dan and Burns, Collin and Basart, Steven and Zou, Andy and Mazeika, Mantas and Song, Dawn and Steinhardt, Jacob},
  title     = {Measuring Massive Multitask Language Understanding},
  booktitle = {Proceedings of the International Conference on Learning Representations (ICLR)},
  year      = {2021}
}

@inproceedings{joshi2017triviaqa,
  author    = {Joshi, Mandar and Choi, Eunsol and Weld, Daniel S. and Zettlemoyer, Luke},
  title     = {{TriviaQA}: A Large Scale Distantly Supervised Challenge Dataset for Reading Comprehension},
  booktitle = {Proceedings of the 55th Annual Meeting of the Association for Computational Linguistics (ACL)},
  pages     = {1601--1611},
  year      = {2017}
}

@article{clark2018think,
  author  = {Clark, Peter and Cowhey, Isaac and Etzioni, Oren and Khot, Tushar and Sabharwal, Ashish and Schoenick, Carissa and Tafjord, Oyvind},
  title   = {Think You Have Solved Question Answering? {Try ARC}, the {AI2} Reasoning Challenge},
  journal = {arXiv preprint arXiv:1803.05457},
  year    = {2018}
}

@inproceedings{kadavath2022language,
  author    = {Kadavath, Saurav and Conerly, Tom and Askell, Amanda and Henighan, Tom and Drain, Dawn and Perez, Ethan and Schiefer, Nicholas and Hatfield-Dodds, Zac and DasSarma, Nova and Tran-Johnson, Eli and others},
  title     = {Language Models (Mostly) Know What They Know},
  booktitle = {arXiv preprint arXiv:2207.05221},
  year      = {2022}
}

@article{lin2022teaching,
  author  = {Lin, Stephanie and Hilton, Jacob and Evans, Owain},
  title   = {Teaching Models to Express Their Uncertainty in Words},
  journal = {Transactions on Machine Learning Research},
  year    = {2022}
}

@inproceedings{kuhn2023semantic,
  author    = {Kuhn, Lorenz and Gal, Yarin and Farquhar, Sebastian},
  title     = {Semantic Uncertainty: Linguistic Invariances for Uncertainty Estimation in Natural Language Generation},
  booktitle = {Proceedings of the International Conference on Learning Representations (ICLR)},
  year      = {2023}
}

@inproceedings{naeini2015obtaining,
  author    = {Naeini, Mahdi Pakdaman and Cooper, Gregory and Hauskrecht, Milos},
  title     = {Obtaining Well Calibrated Probabilities Using {Bayesian} Binning},
  booktitle = {Proceedings of the AAAI Conference on Artificial Intelligence},
  volume    = {29},
  number    = {1},
  year      = {2015}
}

@article{singh2024confidence,
  author  = {Singh, Aniket Kumar and Lamichhane, Bishal and Devkota, Suman and Dhakal, Uttam and Dhakal, Chandra},
  title   = {Do Large Language Models Show Human-like Biases? Exploring Confidence-Competence Gap in {AI}},
  journal = {Information},
  volume  = {15},
  pages   = {92},
  year    = {2024}
}

@article{singh2023cognitive,
  author  = {Singh, Aniket Kumar and Devkota, Suman and Dhakal, Uttam and Dhakal, Chandra},
  title   = {The Confidence-Competence Gap in Large Language Models: A Cognitive Study},
  journal = {arXiv preprint arXiv:2309.16145},
  year    = {2023}
}

@article{singh2025code,
  author  = {Singh, Mukul and Chatterjee, Somya and Radhakrishna, Arjun and Gulwani, Sumit},
  title   = {Do Code Models Suffer from the {Dunning-Kruger} Effect?},
  journal = {arXiv preprint arXiv:2510.05457},
  year    = {2025}
}

@inproceedings{geng2024naacl,
  author    = {Geng, Jiahui and Cai, Fengyu and Wang, Yuxia and Koeppl, Heinz and Nakov, Preslav and Gurevych, Iryna},
  title     = {A Survey of Confidence Estimation and Calibration in Large Language Models},
  booktitle = {Proceedings of the 2024 Conference of the North American Chapter of the Association for Computational Linguistics (NAACL)},
  year      = {2024}
}

@article{xie2024survey,
  author  = {Xie, Liangru and Liu, Hui and Zeng, Jingying and Tang, Xianfeng and Han, Yan and Luo, Chen and Huang, Jing and Li, Zhen and Wang, Suhang and He, Qi},
  title   = {A Survey of Calibration Process for Black-Box {LLMs}},
  journal = {arXiv preprint arXiv:2412.12767},
  year    = {2024}
}

@article{groot2024overconfidence,
  author  = {Groot, Tobias and Valdenegro-Toro, Matias},
  title   = {Overconfidence is Key: Verbalized Uncertainty Evaluation in Large Language and Vision-Language Models},
  journal = {arXiv preprint arXiv:2405.02917},
  year    = {2024}
}

@article{chhikara2025confidence,
  author  = {Chhikara, P.},
  title   = {Mind the Confidence Gap: Overconfidence, Calibration, and Distractor Effects in Large Language Models},
  journal = {arXiv preprint arXiv:2502.11028},
  year    = {2025}
}

@article{leng2024taming,
  author  = {Leng, Jixuan and Huang, Chengsong and Zhu, Banghua and Huang, Jiaxin},
  title   = {Taming Overconfidence in {LLMs}: Reward Calibration in {RLHF}},
  journal = {arXiv preprint arXiv:2410.09724},
  year    = {2024}
}

@article{xiong2023uncertainty,
  author  = {Xiong, Miao and Hu, Zhiyuan and Lu, Xinyang and Li, Yifei and Fu, Jie and He, Junxian and Hooi, Bryan},
  title   = {Can {LLMs} Express Their Uncertainty? An Empirical Evaluation of Confidence Elicitation in {LLMs}},
  journal = {arXiv preprint arXiv:2306.13063},
  year    = {2023}
}

@article{xu2025mirror,
  author  = {Xu, Chenjun and Wen, Bingbing and Han, Bin and Wolfe, Robert and Wang, Lucy Lu and Howe, Bill},
  title   = {Do Language Models Mirror Human Confidence? Exploring Psychological Insights to Address Overconfidence in {LLMs}},
  journal = {arXiv preprint arXiv:2506.00582},
  year    = {2025}
}

@article{steyvers2024llms,
  author  = {Steyvers, M. and Tejeda Lemus, Heliodoro and Kumar, Aakriti and Bel{\'e}m, Catarina G. and Karny, Sheer and Hu, Xinyue and Mayer, Lukas William and Smyth, P.},
  title   = {What Large Language Models Know and What People Think They Know},
  journal = {Nature Machine Intelligence},
  volume  = {7},
  pages   = {221--231},
  year    = {2024}
}

@article{yoon2025reasoning,
  author  = {Yoon, Dongkeun and Kim, Seungone and Yang, Sohee and Kim, SunKyoung and Kim, Soyeon and Kim, Yongil and Choi, Eunbi and Kim, Yireun and Seo, Minjoon},
  title   = {Reasoning Models Better Express Their Confidence},
  journal = {arXiv preprint arXiv:2505.14489},
  year    = {2025}
}

@article{pawitan2024confidence,
  author  = {Pawitan, Yudi and Holmes, Chris},
  title   = {Confidence in the Reasoning of Large Language Models},
  journal = {arXiv preprint arXiv:2412.15296},
  year    = {2024}
}

@inproceedings{zhu2023calibration,
  author    = {Zhu, Chiwei and Xu, Benfeng and Wang, Quan and Zhang, Yongdong and Mao, Zhendong},
  title     = {On the Calibration of Large Language Models and Alignment},
  booktitle = {Proceedings of the 2023 Conference on Empirical Methods in Natural Language Processing (EMNLP)},
  year      = {2023}
}

@article{park2026finetuning,
  author  = {Park, Sangjun and Meyerson, Elliot and Qiu, Xin and Miikkulainen, Risto},
  title   = {Fine-Tuning Language Models to Know What They Know},
  journal = {arXiv preprint arXiv:2602.02605},
  year    = {2026}
}

@article{zhang2023hallucination,
  author  = {Zhang, Yue and Li, Yafu and Cui, Leyang and Cai, Deng and Liu, Lemao and Fu, Tingchen and Huang, Xinting and Zhao, Enbo and Zhang, Yu and Chen, Yulong and Wang, Longyue and Luu, A. and Bi, Wei and Shi, Freda and Shi, Shuming},
  title   = {Siren's Song in the {AI} Ocean: A Survey on Hallucination in Large Language Models},
  journal = {arXiv preprint arXiv:2309.01219},
  year    = {2023}
}

\end{document}